\pdfoutput=1

\documentclass[11pt]{article}

\usepackage[final]{ACL2023}

\usepackage{times}
\usepackage{latexsym}
\usepackage{amsmath}
\usepackage{graphicx}
\usepackage{microtype}
\usepackage{algorithm}
\usepackage{algorithmic}
\usepackage{multirow}
\usepackage{booktabs}

\usepackage[T1]{fontenc}

\usepackage[utf8]{inputenc}

\usepackage{microtype}

\usepackage{inconsolata}

\title{GraphERE: Jointly Multiple Event-Event Relation Extraction via Graph-Enhanced Event Embeddings}

\author{Haochen Li \\
  Peking University, Beijing, China \\
  \texttt{haochenli@pku.edu.cn} \\\And
  Di Geng \\
  Peking University, Beijing, China \\
  \texttt{gengdi@stu.pku.edu.cn} \\}

\begin{document}
\maketitle
\begin{abstract}
Events describe the state changes of entities. In a document, multiple events are connected by various relations (e.g., Coreference, Temporal, Causal, and Subevent). Therefore, obtaining the connections between events through Event-Event Relation Extraction (ERE) is critical to understand natural language.
There are two main problems in the current ERE works: a. Only embeddings of the event triggers are used for event feature representation, ignoring event arguments (e.g., time, place, person, etc.) and their structure within the event. b. The interconnection between relations (e.g., temporal and causal relations usually interact with each other ) is ignored.
To solve the above problems, this paper proposes a jointly multiple ERE framework called GraphERE based on Graph-enhanced Event Embeddings. First, we enrich the event embeddings with event argument and structure features by using static AMR graphs and IE graphs; Then, to jointly extract multiple event relations, we use Node Transformer and construct Task-specific Dynamic Event Graphs for each type of relation. Finally, we used a multi-task learning strategy to train the whole framework.
Experimental results on the latest MAVEN-ERE dataset validate that GraphERE significantly outperforms existing methods. Further analyses indicate the effectiveness of the graph-enhanced event embeddings and the joint extraction strategy.

\end{abstract}

\begin{figure}[h]
  \includegraphics[width=8cm]{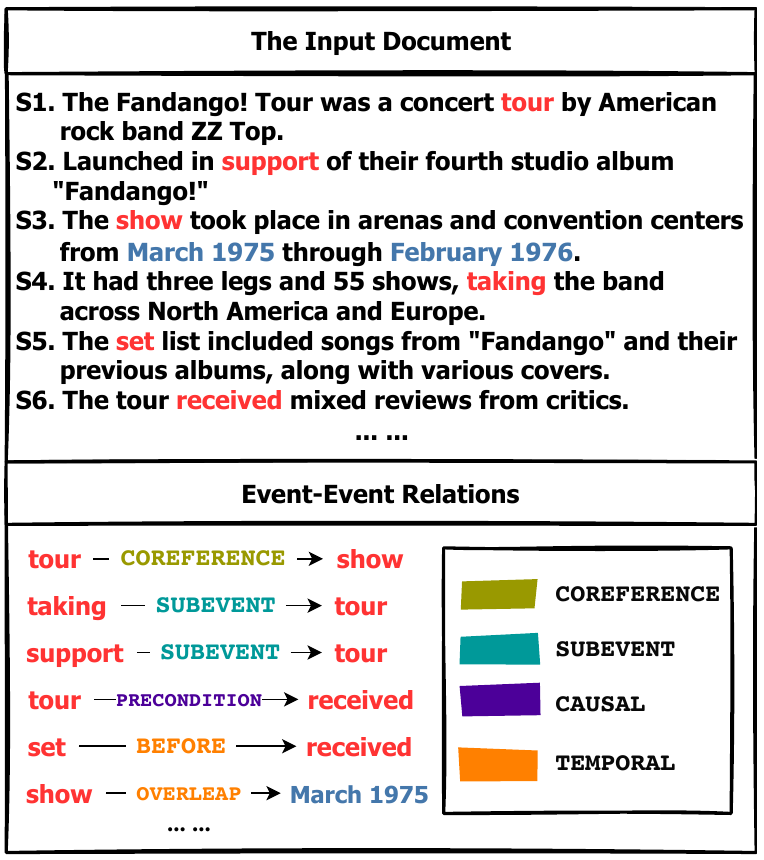}
  \caption{An example of the ERE task.}
  \label{fig:1}
\end{figure}

\section{Introduction}
Event-Event Relation Extraction (ERE) is an essential task in Information Extraction that aims to identify the semantic relationships among different events in the text. 
There are various and complex types of event relations in the real world. Figure \ref{fig:1} illustrates an example document, the event 
\underline{\textit{tour} }\textbf{(S1)} is the \textbf{precondition} of the event \underline{\textit{received}} \textbf{(S6)} and the event 
\underline{\textit{set}} \textbf{(S5)} happens \textbf{before} the event \underline{\textit{received}} \textbf{(S6)}. 
ERE methods should be able to recognize the causal (\textbf{precondition}) and temporal (\textbf{before}) relations given the two events.
With the development of Event Extraction (EE) and Event Knowledge Graph (EKG) construction, ERE has become essential to link extracted events to get EKGs \citep{li_future_2021,zhang_eventke_2021}. 
In addition, ERE also has applications in many other downstream NLP tasks such as Automatic Summary and Machine Q\&A \citep{chaturvedi_story_2017,huang_cosmos_2019,ning_torque_2020}.

Early statistical ERE methods \citep{chambers_classifying_2007,zhao_event_2016} use manually designed features, and neural network-based approaches \citep{cheng_classifying_2017,li_knowledge-oriented_2019} have been utilized to learn hidden features for extraction automatically. Recently, Pretrained Language Models (PLM), such as BERT \cite{devlin_bert_2019}, have gained more attention for the excellent performance on many NLP tasks. Some studies adopt PLM to encode the text and incorporate external knowledge to enhance the event embeddings \citep{liu_knowledge_2020,zuo_knowdis_2020}. 

One major limitation of current methods is that they ignore the importance of event argument and structure information, only using event triggers' token embeddings as event representation. As it is shown in Figure \ref{fig:1}: the \underline{\textit{tour} }\textbf{(S1)} event has the argument \textit{American rock band ZZ Top}. While the \underline{\textit{support} }\textbf{(S2)} event has the argument \textit{studio album}. A rock band is likely to release an album so that the \underline{\textit{support} }\textbf{(S2)} event may have relations to the \underline{\textit{tour} }\textbf{(S1)} event. Further, events having similar arguments (e.g., time, location, and subject) tend to be related. Thus event argument and structure information are critical to event relations.
To address the problem, we adopt NLP tools, including AMR(\citeauthor{banarescu2013abstract}) and OpenIE(\citeauthor{angeli2015leveraging}) to extract event arguments automatically and build Static Event Graphs. In such graphs, event triggers and arguments are the nodes, and they are connected by a subordinate relationship. Then we use a Graph Encoder such as Graph Attention Network (GAT, \citeauthor{velickovic2017graph}) to get event embeddings with argument and structure features.


Another limitation is that current methods only consider one or two event relation types. Such as \citealt{zhang_knowledgedata_2022,li_document-level_2022} only focus on event temporal relation; \citealt{liu_knowledge_2020} adds external knowledge only to extract causal relations. However, event relations interact with each other. For example in Figure \ref{fig:1}, the event \underline{\textit{tour} }\textbf{(S1)} is coreference with the event \underline{\textit{show} }\textbf{(S3)}. Meanwhile, the event \underline{\textit{show} }\textbf{(S3)} is have a temporal overlap relation with the timex (words represent time) \underline{\textit{March 1975}}, so the \underline{\textit{tour} }\textbf{(S1)} event may also have temporal relation with \underline{\textit{March 1975}}. 
To utilize the interactions, we extract four event relations (Coreference, Temporal, Causal, and Subevent) simultaneously and train a joint model using a multi-task learning strategy.

\begin{figure*}[h]
  \centering
  \includegraphics[width=16cm]{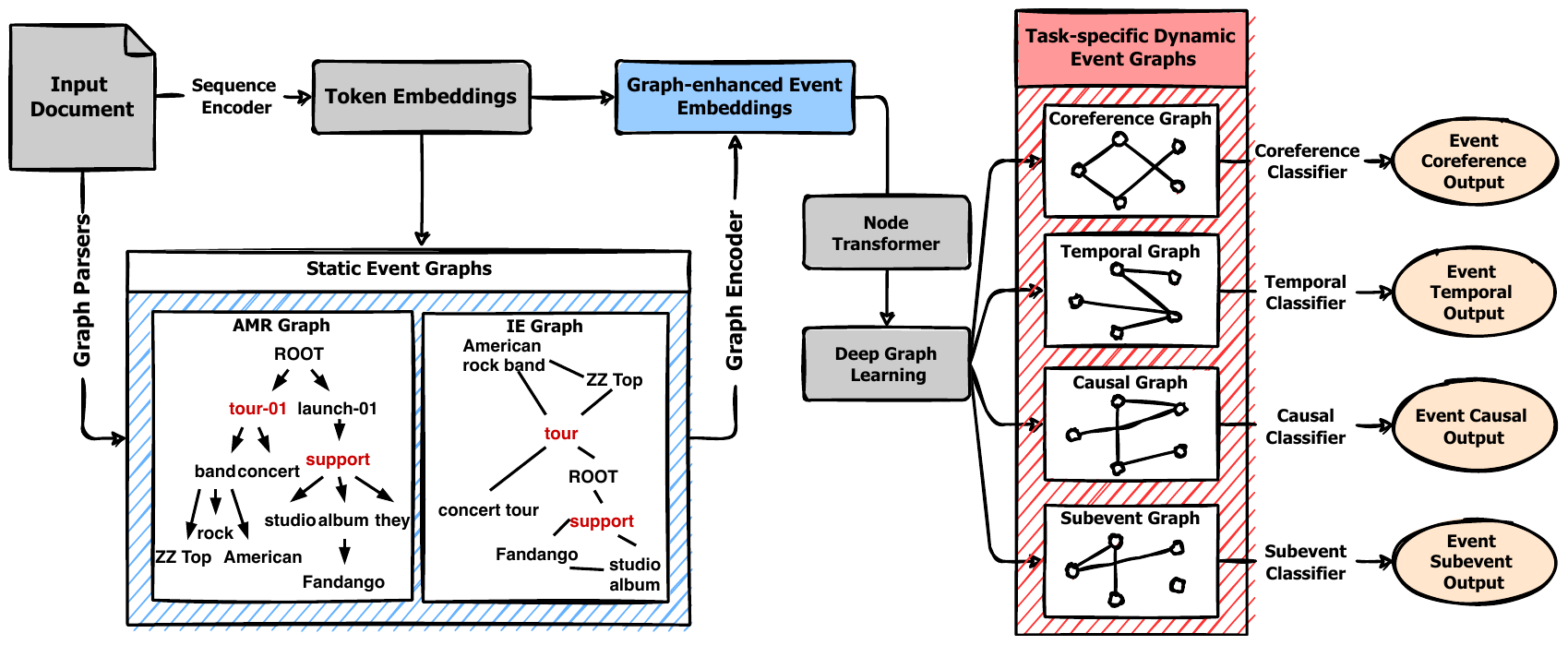}
  \caption{The overall architecture of GraphERE.}
  \label{fig:2}
\end{figure*}

Specifically, we propose a jointly multiple event-event relation extraction framework called GraphERE (shown in Figure \ref{fig:2}). 
Given a document, firstly, we use a \textbf{Sequence Encoder} to get the initial token embeddings. 
Secondly, we apply NLP parsers to acquire \textbf{Static Event Graphs} including an AMR Graph and an IE Graph, and use a graph encoder to get \textbf{Graph-enhanced Event Embeddings}.
Thirdly, we use \textbf{Node Transformer} and \textbf{Deep Graph Learning} algorithms to construct \textbf{Task-specific Dynamic Event Graphs}. Each dynamic event graph is related to an event relation type followed by the corresponding classifier.
Finally, we use a multi-tasking learning approach to jointly train on the four ERE subtasks and optimize the parameters of the whole framework.

Our main contributions are as follows:
\begin{itemize}
\item Graph-enhanced Event Embeddings are proposed to encode features of event argument and structure. To achieve this, we construct Static Event Graphs containing event triggers, arguments, and their subordination.
\item A framework for jointly multiple event-event relation extraction is proposed (GraphERE). We introduce Node Transformer and Task-specific Dynamic Graphs to model the feature interaction of four event relation types ( Coreference, Temporal, Causal, and Subevent).
\item Experimental results on the MAVEN-ERE dataset show that GraphERE significantly overpasses the baselines. Further analyses also indicate the effectiveness of the graph-enhanced event embeddings and the joint extraction strategy.

\end{itemize}

\section{Related Work}
\subsection{Event-Event Relation Extraction}
Due to the diversity of event relations, most previous research works only focus on one type of event relation. 
\textbf{Temporal Relation}: \citealt{li_document-level_2022} selects the most relevant contextual sentences for document-level event temporal relation extraction. \citealt{mathur_doctime_2022} presents a novel temporal dependency graph parser integrating longer term dependencies with a novel path prediction loss. 
\citealt{wen_utilizing_2021} introduces relative time, which is an actual number indicating the relative position of the event in the timeline to solve event-event temporal relation classiﬁcation.
\textbf{Causal Relation}: \citealt{liu_knowledge_2020} designs a knowledge-aware causal reasoner and a mention masking reasoner to exploit an external knowledge base \textit{ConceptNet} and alleviate its incompleteness. \citealt{zuo_learnda_2021} introduces a new approach to augment training data for event causality identiﬁcation, by iteratively generating new examples and classifying event causality in a dual learning framework.
\textbf{Coreference Relation}: \citealt{huang_improving_2019} proposes to transfer argument compatibility knowledge to the event coreference resolution system and iteratively uses the coreference model to relabel the argument compatibility instances.\citealt{minh_tran_exploiting_2021} introduces document structures to improve representation learning and proposes to exploit the consistencies between golden and predicted clusters of event mentions.
\textbf{Subevent Relation}: \citealt{aldawsari_detecting_2019} incorporates several novel discourse and narrative features to improve identifying the internal structure of events.

Recently, some studies extracting two types of event relations are also inspiring, which can exploit the interaction between different relations.
\citealt{wang_joint_2020} enforces a comprehensive set of logical constraints to connect temporal and subevent relation extraction tasks. \citealt{hwang_event-event_2022} proposes to represent each event as a probabilistic box, which can handle antisymmetry between events.
\citealt{man_selecting_2022} proposes identifying the most relevant contextual sentences, which can address the length limitation in existing transformer-based models, to extract subevent and temporal relations. 
To take full advantage of more relation interaction and address the problems of lacking large-scale, high-quality datasets, \citealt{wang_maven-ere_2022} constructs a unified large-scale annotated ERE dataset ( MAVEN-ERE) and proposes a RoBERTa baseline based on it.

\subsection{Event Graph Construction}
Event graphs can complement dynamic and procedural knowledge to the original entity-based knowledge graph. Meanwhile, encoding event graphs can get graph-based embeddings for entity and event, which are helpful in downstream tasks like event and event relation extraction.
\citealt{zeng_gene_nodate} encodes event graphs by introducing a multi-view graph encoder. 
\citealt{du_learning_2021} proposes a variational autoencoder-based model, which employs a latent variable to capture the commonsense knowledge from the event graph.
\citealt{du_excar_2021} proposes a graph-based explainable causal reasoning framework by using the logical law behind the causality.
\citealt{li_future_2021} captures and organizes related events in graphs via their coreferential or related arguments for event schema induction.
\citealt{yuan_event_2021} proposes to construct an event graph to capture the relations and structures between events and use it to instruct sentence fusion.
\citealt{zhang_eventke_2021} represents an event-enhanced knowledge graph, adding event nodes to the original KG as a heterogeneous network. 
\citealt{tran_phu_graph_2021} constructs interaction graphs to capture relations between events for event causality identification.

\section{Approach}
\subsection{Task Formulation}
This paper focuses on the document-level Event-Event Relation Extraction (ERE) task. The input Document $D$ contains $m$ sentences $[s_1, s_2, ..., s_m]$, and also can be seen as a sequence of $n$ tokens $X = [x_1, x_2, ..., x_n]$. Each document contains $p$ known event triggers (the most representative words for each event) $E = [e_1, e_2, ..., e_p]$ and $q$ timexes (words indicating time)  $T=[t_1, t_2, ..., t_q]$. Each event trigger or timex is represented by specific tokens in the document, thus $E, T \subseteq X$.

This paper focuses on the four types of most widely studied event relations: Coreference, Temporal, Causal, and Subevent, and each type of relation has several subtypes (e.g., BEFORE, AFTER, and so on in Temporal). The ERE task aims to extract the relation $R$ between event-event pairs and event-timex pairs (only in the Temporal relation):
\begin{equation}
    \begin{aligned}
    R=&[<e_i, e_j,r_{ij}>, <e_i,t_k,r_{ik}>],\\
    &1 \le i, j \le p, 1 \le k \le q
    \end{aligned}
\end{equation}

\subsection{Sequence Encoder}
We first use a pre-trained language model (RoBERTa \cite{liu2019roberta} in this paper) as the sequence encoder to encode the input :
\begin{equation}
    h_1, h_2, ... , h_n = RoBERTa([x_1, x_2, ..., x_n]), 
\end{equation}

Where $h_i$ is the embedding of the token $x_i$. If the document's length exceeds the input length of the language model, we will segment a document into multiple shorter input sequences.

Each event trigger $e_i$ in the document may likely be divided into multiple sub-words by the encoder. We average the embeddings of the sub-words to get the embedding. Then, we can get the initial token embeddings for all the events:
\begin{equation}
        H_{Event} = [h_{e_1}, h_{e_2},..., h_{e_p}]
\end{equation}

\begin{figure*}[h]
  \centering
  \includegraphics[width=16cm]{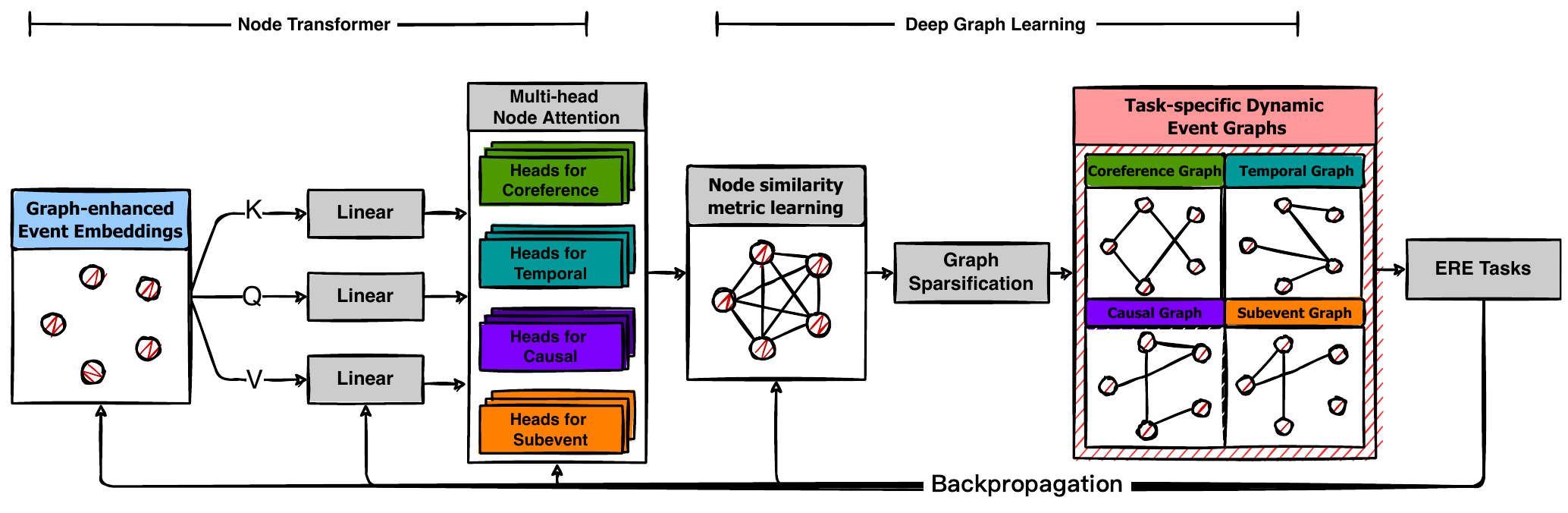}
  \caption{Procedures for the Node Transformer and Deep Graph Learning algorithms.}
  \label{fig:3}
\end{figure*}

\subsection{Graph-enhanced Event Embeddings by Static Event Graphs}
Sequence Encoder can obtain the semantic and contextual features of the event trigger. However, each event contains several event arguments (entities that participat in the event). Also, there is a graph structure between event triggers and arguments, e.g., a single event is a subgraph in which an event trigger and several event arguments are connected, while different events may contain same arguments and thus are indirectly connected. Adding the features of event argument and structure to event embeddings can better encode events, benefiting the downstream event relation extraction.

To obtain event arguments and event graph structures, we construct graph parsers through existing research works and build Static Event Graphs for each input document. For the same document, the nodes and edges of the static event graphs are fixed and will not change with the training process. In this paper, we choose the AMR Parser and OpenIE parser to construct AMR Graph $G_{AMR}$ and IE graph $G_{IE}$, respectively.

\textbf{AMR Graph:} Abstract Meaning Representation (AMR, \citeauthor{banarescu2013abstract}) is a way to obtain the semantic structure of the text. We use a pretrained AMR parser\footnote{https://github.com/bjascob/amrlib} to parse the input document into AMR structures. Each AMR structure is a directed acyclic graph with concepts as nodes and semantic relations as edges. This structure is similar to the form of events and thus can be used to extract the arguments and structural features of events in the text without supervision. With the AMR parser, we can obtain a static AMR graph $G_{AMR}=<V_{AMR}, C_{AMR}>$, where $V_{AMR}$ represents the nodes and $C_{AMR}$ represents the edges in the graph.

\textbf{IE Graph:} The Information Extraction (IE) task aims to extract information such as entities and entity relationships from natural language. We want to represent the IE result in a graph structure to represent high-level information in the document. We use OpenIE parser\footnote{https://stanfordnlp.github.io/CoreNLP/openie.html} to parse the input document into multiple entity triples <head entity, tail entity, relationship>, e.g., \textit{<American rock band, ZZ Top, name>}. We construct a graph by taking all the head and tail entities of the triples as nodes and their relations as edges. Then we can get a static IE graph $G_{IE}=<V_{IE}, C_{IE}>$, where $V_{IE}$ represents the nodes and $C_{IE}$ represents the edges.

After obtaining the static event graphs, we map the nodes in the graph to the events in the document and then encode the information using a graph encoder to enhance the event embeddings. We use the event embeddings in Section 3.2 as the initial feature of the nodes in the graph and Graph Attention Networks (\citeauthor{velickovic2017graph}) as the graph encoder to capture the events’ argument and structural features. Taking the AMR graph as an example. If the node $v_i\in V_{AMR}$ in $G_{AMR}$ has a neighbor node $v_j\in V_{AMR}$. First, we calculate the attention coefficient $co_{ij}$ between nodes $v_i$ and $v_j$ by using their embeddings.
\begin{equation}
	co_{ij}=< W^{amr}h_{v_i}, W^{amr}h_{v_j} > ,j\in  N(e_i)
\end{equation}
Where $W^{amr}$ is a transformation matrix for the AMR graph, $h_{v_i}$ stands for the initial embedding of node $i$, $< \cdot , \cdot > $ stands for the inner product operation, and $N({e_i})$ is the set of all neighbors of node $i$. 
Then we normalize the attention coefficients between node $v_i$ and all neighbors.
\begin{equation}
	\alpha_{ij} =\frac{exp\left ( LeakyReLU\left ( co_{ij}  \right )  \right ) }{ {\textstyle \sum_{k\in N(e_i)}^{}}exp\left ( LeakyReLU\left ( co_{ik}  \right )  \right )  }
\end{equation}
$LeakyReLU$ is the activation function. After normalization, we can obtain the AMR graph-based embeddings of each node.
\begin{equation}
	h_{e_i}^{amr} =\sigma \left (  {\textstyle \sum_{j\in N({e_i}) }^{}} \alpha _{ij} W^{amr}h_{v_j}  \right )
\end{equation}
Where $\sigma$ stands for the activation function.
Similarly, we can obtain the node embeddings $h_{v_i}^{ie}$ in the IE graph. Then we add the embedding of the corresponding event node in the static event-based graphs and its initial token embedding, and we can obtain graph-enhanced event embeddings $gh$.
\begin{equation}
    gh_{e_i} = h_{e_i} + \beta h_{e_i}^{amr} + (1-\beta) h_{e_i}^{ie}
\label{eq:7}
\end{equation}

Where $\beta$ is the mix ratio and $e_i\in E$.

\subsection{Joint Multiple Extraction by Task-specific Dynamic Event Graphs}
To capture interactions within multiple types of event relation, we employ a multi-head attention module called Node Transformer. Then we construct four Task-specific Dynamic Event Graphs using a Deep Graph Learning algorithm (shown in Figure \ref{fig:3}) to extract each event relation separately. The details of this section are explained below.

\textbf{Node Transformer:} We use a self-attention mechanism (\citeauthor{vaswani2017attention}) for that each event node can pass features with other nodes. the graph-enhanced event embeddings $gh$ are used to generate the $K(keys)$, $Q(queries)$ and $V(values)$ in the attention layer. Further, we extend the attention layer to a multi-head version for capturing more event semantics.
\begin{equation}
    Attention(gh) = softmax(\frac{gh{gh}^T}{\sqrt{d_h}})gh
\end{equation}
\begin{equation}
    \Tilde{gh} = concat(Att_1, ...,Att_h)W
\end{equation}
Where $d_h$ stands for the dimension of event embedding, the $Att_i$ represents each head's attention function, the $head$ represents the number of head and the $W$ is a learnable parameter. We use $4\times head$ to calculate the semantics of four event relations and provide event node embeddings for downstream task-specific dynamic event graphs construction.

\textbf{Deep Graph Learning:} The embeddings of event node are then refined through a Deep Graph Learning algorithm, where we construct a task-specific dynamic event graphs for each relation type. Unlike static event graphs, which are fixed, the connection of dynamic graphs will change concerning the event embeddings. The ultimate goal is to learn the optimized graph structure and get the refined event node embeddings for each downstream relation classification task.

To construct a dynamic graph from the event embeddings, inspired by \citealt{chen_iterative_2020}, we first use a graph metric function( weighted cosine in our work) to compute the pair-wise event node similarity and return a fully-connected weighted graph. We use four weight vectors $w_r, r \in [coreference, temporal, causal, subevent]$ to compute four independent similarity matrices $S_r$ as the initial graphs of every event relation.
\begin{equation}
    s_{ij}^r = cos(w_r \circ \Tilde{gh}_{e_i} w_r \circ \Tilde{gh}_{e_j}) 
\end{equation}
Where $\circ$ represents the Hadamard product, and $w_r$s are learnable parameters which have the same dimension as the input event embeddings $\Tilde{gh}_{e_i}$ and $\Tilde{gh}_{e_j}$.  $s_{ij}^r$ computes the cosine similarity between the two input vectors for the $r$-th type of event relation semantics captured from the embeddings. 

As the initial graph is fully connected, it has a high computational cost and introduces a lot of noise. We propose Graph Sparsification to generate a sparser non-negative dynamic graph $\mathcal{A}$ from the initial graph $\mathcal{S}$. Specifically, we use thresholds $\epsilon_r$ to mask the edges which have smaller weights.
\begin{equation}
mask^r=\left\{
\begin{array}{rcl}
0 & & {s_{ij}^r <= \epsilon_r}\\
1 & & {s_{ij}^r > \epsilon_r}\\
\end{array} \right.
\label{eq:11}
\end{equation}
\begin{equation}
\mathcal{A}^r=mask^r*\mathcal{S}^r
\end{equation}
where the $\epsilon_r$ denotes task-specific threshold for Coreference, Temporal, Causal and Subevent relations.
Graph Convolutional Networks (GCN, \citeauthor{kipf2016semi}) are then adopted for each dynamic graph, which take the event node feature matrix $\Tilde{gh}$ and adjacency matrix $\mathcal{A}^r$ as inputs to compute the refined event embeddings.
\begin{equation}
\overline{gh_{e_i}^r} = \sigma(\sum_{j\in\mathcal{A}^r(i)}\frac{1}{c_{ij}}W^r\Tilde{gh_{e_j}}^{(l)} + b^r)
\end{equation}
Where $\mathcal{A}(i)$ is the set of neighbors of node $i$, $c_{ij}$ is the product of the square root of node degrees
(i.e.,$c_{ij} = \sqrt{|\mathcal{A}(i)|}\sqrt{|\mathcal{A}(j)|})$ and $\sigma$ is an activation function. $W^{r}$ and $b^{r}$ are learnable parameters.

Then the probability distribution of the relation between event pairs can be obtained by task-specific linear classifiers.
\begin{equation}
    \hat{R}_r = softmax(W_r^{linear}\overline{gh^r}+b_r^{linear})
\end{equation}

\subsection{Training GraphERE}
For each event relation, we obtained the predicted probability distribution by using the classifier in 3.4. We use the cross-entropy function to calculate the loss between the predicted results and the golden annotation.
\begin{equation}
loss_r = crossentropy(\hat{R}_r, R_r)
\label{eq:15}
\end{equation}
Where $\hat{R}_r$ and $R_r$ stand for the predicted and golden event relations, respectively. 
We use a multi-task learning strategy by adding the loss of all event relations and optimizing all the parameters through joint training to achieve jointly multiple event relation extraction.
\begin{equation}
loss = \sum\lambda_rloss_r
\label{eq:16}
\end{equation}
Where $\lambda_r$ denotes the different weight of each event relation when added to the total $loss$. The parameters in the Sequence Encoder, Static Graph Encoder, Node Transformer, Deep Graph Learning, and task-specific classifiers will be optimized during the training process. 
\section{Experiments}
\begin{table*}[]
\centering
\small
\begin{tabular}{@{}l|ccc|ccc|ccc|ccc@{}}
\toprule
\multirow{2}{*}{Model} & \multicolumn{3}{c|}{Coreference}                                        & \multicolumn{3}{c|}{Temporal}                                           & \multicolumn{3}{c|}{Causal}                                             & \multicolumn{3}{c}{SubEvent}                                           \\ \cmidrule(l){2-13} 
                       & \multicolumn{1}{c|}{P} & \multicolumn{1}{c|}{R} & F1                    & \multicolumn{1}{c|}{P} & \multicolumn{1}{c|}{R} & F1                    & \multicolumn{1}{c|}{P} & \multicolumn{1}{c|}{R} & F1                    & \multicolumn{1}{c|}{P} & \multicolumn{1}{c|}{R} & F1                   \\ \midrule
Interact                                       & 74.62       & \textbf{84.71}            & 79.67                 & -                      & -                      & -                     & -                      & -                      & -                     & -                      & -                      & -                    \\
DocTime                & -                      & -                      & -                     & 54.40                  & 52.97                  & 53.68                 & -                      & -                      & -                     & -                      & -                      & -                    \\
Liu's knowledge                      & -                      & -                      & -                     & -                      & -                      & -                     & 28.52                  & 30.16                  & 29.34                 & -                      & -                      & -                    \\
Aldawsari's feature                      & -                      & -                      & -                     & -                      & -                      & -                     & -                      & -                      & -                     & 16.35                  & 19.84                  & 18.10                \\
RoBERTa$_{split}$        & 78.14                  & 81.64                  & 79.87                 & 53.05                  & 53.29                  & 53.17                                  & 28.10        & 32.15           & 29.59                  & 27.34                  & 20.91                  & 23.69               \\
RoBERTa$_{joint}$        & 81.65                  & 77.76                  & 79.39                 & 50.73                  & \textbf{56.73}                  & 53.56                                   & 27.97   & 33.28               & 30.39  & 19.90                  & 24.17                  & 21.83                               \\ \midrule
GraphERE$_{split}$       & 81.69                  & 77.66                  & 79.62               & \textbf{57.09}   & 50.02                                   & 53.32                 & 32.99                  & 28.19                  & 30.40                 & 26.10                  & 20.79                  & 22.08                \\
GraphERE$_{joint}$       & \textbf{82.20}                  & 78.17                  & \textbf{80.13}                 & 55.05                  & 54.43                  & \textbf{54.74}                 & 28.99                  & \textbf{33.42}                  & \textbf{31.05}                 & \textbf{31.23}                  & \textbf{24.23}                  & \textbf{27.26}                \\ \midrule\midrule
w/o Dynamic Graphs     & 81.85	&77.27	&79.50	&53.18	&53.25	&53.22	&\textbf{33.39}	&26.75	&29.70	&30.05	&18.12	&22.61   \\
w/o Static Graphs      & 77.84	&77.53	&77.68	&52.99	&51.56	&52.27	&30.82	&28.30	&29.50	&28.49	&19.84	&23.39 \\
w/o Node Transformer   & 82.03	&75.34	&79.84	&53.89	&54.22	&54.05	&32.83	&27.21	&29.76	&30.38	&21.84	&25.41 \\ 
\bottomrule
\end{tabular}
\caption{The result for the overall experiment and ablation analysis, and 'w/o' stands for the removal of the module. All the experiments are repeated 10 times, and the average results are recorded.}
\label{tab:1}
\end{table*}
\subsection{Dataset and Setup}
\textbf{Dataset.} We evaluate our model on the latest MAVEN-ERE dataset\citep{wang_maven-ere_2022}, containing 4,480 documents and 103,193 events annotated with 4 relation types and 10 subtypes. For fair competition, we follow the same preprocessing step as the baselines\footnote{https://github.com/THU-KEG/MAVEN-ERE}. \\
\textbf{Setup.} \textit{Roberta-base} is used in the sequence encoder module. We use 0.8 as the mix ratio $\beta$ (eq.\ref{eq:7}) to incorporate the AMR and IE graphs. We use 4 heads with a dropout rate of 0.3 in Node Transformer. The non-negative threshold $\epsilon$ (eq.\ref{eq:11}) in Graph Sparsification and coefficient $\lambda$ (eq.\ref{eq:16}) in Task-specific Classifier are different for each relation extraction task. We use $\epsilon_r$=0.6, 0.4, 0.6, 0.8 and $\lambda_r$=0.5, 1.0, 5.0, 5.0 for Coreference, Temporal, Casual, and Subevent relation extraction, respectively. We optimize our model with AdamW for 30 training epochs and 8 as the batch size, with a learning rate of 2e-5 for RoBERTa and a learning rate of 5e-4 for other parameters. The experiments are conducted on a single NVIDIA GeForce RTX A6000. To speed up the training process, we first obtained and cached the static event graphs of each document. Each batch of training costs about 4 minutes and each epoch costs about 2 hours.\\
\textbf{Metrics.} We adopt MUC Precision, Recall and F1 \citep{vilain_model-theoretic_1995} for event coreference. For the other relation extraction tasks, we use the standard micro-averaged Precision, Recall and F1 Scores.\\
\textbf{Baselines.} Our main baseline is simple RoBERTa model provided by \citealt{wang_maven-ere_2022}, which only encode the whole document with RoBERTa obtaining event embeddings for pair-wise classification. According to whether multiple event relations are trained simultaneously, we set RoBERTa$_{split}$ and RoBERTa$_{joint}$ as comparisons.
In addition, for each relation extraction task, we include related baseline in our experiments. 
\textbf{Interact} \citep{huang_improving_2019} addresses coreference resolution in an iterative way. \textbf{DocTime} \citep{mathur_doctime_2022} pays attention to temporal dependency graph construction. \textbf{Liu's knowledge-enhanced} \citep{liu_knowledge_2020} proposes to incorporate external knowledge from \textit{ConceptNet}for event causality identification. \textbf{Aldawsari's feature-enhanced} \citep{aldawsari_detecting_2019} incorporates discourse and narriative features to extract subevents.

\subsection{Overall Results}
The results of the overall experiment are shown in Table \ref{tab:1}. The proposed GraphERE$_{joint}$ model outperforms all baselines in F1 scores for the four event relations. Compared with the best existing methods for each subtask, our method improves by 0.26\% (Coreference), 1.06\% (Temporal), 0.66\% (Causal), and 3.57\% (Subevent). Further analyses are as follows.\\
(1) GraphERE has a very considerable improvement in Precision. GraphERE$_{joint}$ improves on average 1.39\% over the best baseline on the four tasks (GraphERE$_{split}$ for 1.49\%). This indicates that we introduces prior knowledge from other NLP tools in graph-enhanced event embeddings, which enriches the semantic features in event embeddings and improves the performance in Precision. However, due to the limited extraction capability of these NLP tools, there will be cases where the event arguments and event structures cannot be extracted, so the recall performance of the model is improved relatively little.\\
(2) Comparing RoBERTa$_{joint}$ with RoBERTa$_{split}$, the results are better on the temporal and causal tasks, however, the performance decreases on the remaining two tasks. In contrast, GraphERE$_{joint}$ improved results on all four event relations from GraphERE$_{split}$. Consider that the Roberta baseline uses the same event embeddings to connect multiple classifiers for multi-task learning. During joint training, event embeddings will be traded off among multiple tasks, sacrificing performance on some tasks. In contrast, GraphERE$_{joint}$ uses task-specific dynamic graphs for joint training. The dynamic graphs for each event relation are optimized through training, thus preventing the event embeddings from being interferenced by a particular relation.\\
\begin{figure*}[h]
\centering
  \includegraphics[width=16.5cm]{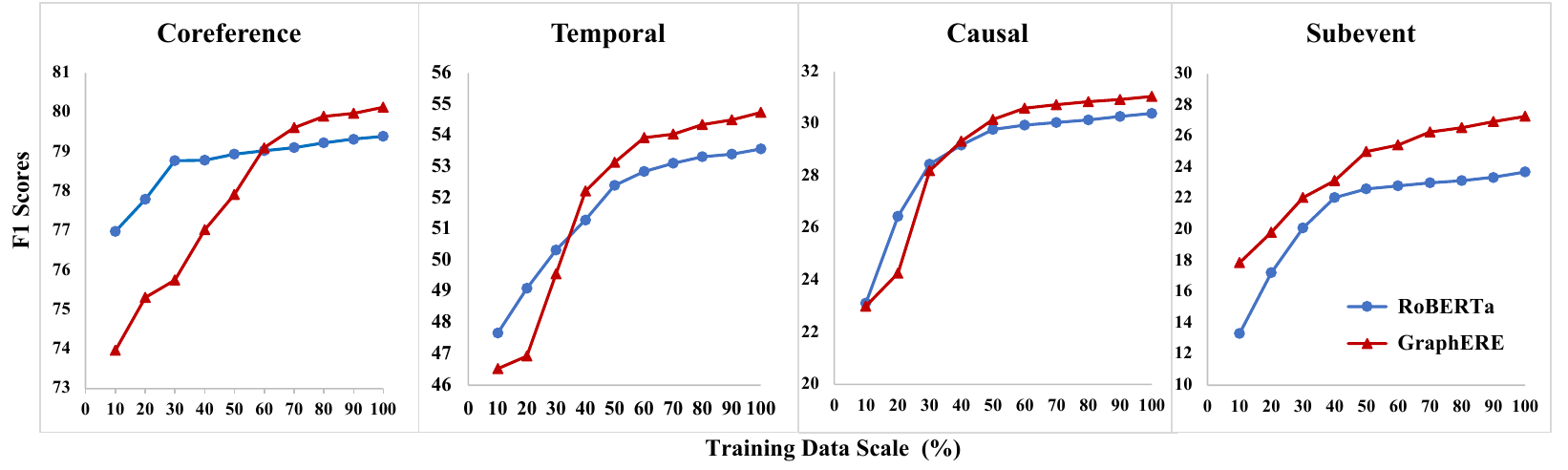}
  \caption{The performance along with the training data scale. We compare our GraphERE$_{joint}$ with the RoBERTa$_{joint}$ baseline, and the average F1 scores of 10 experiments are used.}
  \label{fig:4}
\end{figure*}
(3) The subevent relation is the most challenging task in the four tasks, with all the models exhibiting less than 30\% in F1 Score. However, RoBERTa$_{joint}$ has the most apparent advantage of 3.57\% in the subevent task over baselines. Meanwhile, the improvement from GraphERE$_{split}$ to GraphERE$_{joint}$ is also the largest,  up to 5.18\%. The possible reason is that the subevent relationship has to consider other relations between events (e.g., coreference and temporal order) in addition to the features within events, which makes our joint training strategy make the most gains.

\subsection{Ablation Analysis}
To verify the effectiveness of each part of GraphERE, we conducts ablation analysis experiments. The results are shown in Table \ref{tab:1}, from which it can be found: \\
(1) After removing the dynamic graphs, the model's ability to capture the information interactions between multiple event relations also decreases, so it performs poorly on the more difficult tasks of Causal and Subevent (1.35\% and 4.65\% in F1 scores, respectively.). Meanwhile, the model's Recall performance drops significantly by an average of 3.72\% on the four relations. \\
(2) After removing the static graphs, the Precision performance of the model declines because of the missing features of event argument and structure, with an average decline of 1.83\% on the four tasks. \\
(3) After removing Node Transformer, the model shows a decrease in performance on all four relations, with an average decrease of 1.03\% in F1 Scores. This indicates the importance of information interaction between event nodes

\begin{table}[]
\centering
\small
\begin{tabular}{@{}c|cccc@{}}
\toprule
$\beta$ & Coreference & Temporal & Causal & Subevent \\ \midrule
0.0 & 77.64       & 50.41    & 29.14  & 22.11    \\
0.1 & 78.53       & 51.42    & 30.04  & 23.35    \\
0.2 & 78.63       & 51.53    & 30.27  & 23.75    \\
0.3 & 78.81       & 51.78    & 30.62  & 23.99    \\
0.4 & 79.48       & 51.68    & 30.29  & 24.03    \\
0.5 & 78.99       & 52.01    & 29.58  & 24.27    \\
0.6 & 79.32       & 52.77    & 30.74  & 24.65    \\
0.7 & \textbf{80.42}       & 52.98    & 30.44  & 25.51    \\
0.8 & 80.13       & \textbf{54.74}    & 31.05  & \textbf{26.26}    \\
0.9 & 79.57       & 53.83    & 31.18  & 25.22    \\
1.0 & 79.53       & 53.5     & \textbf{31.46}  & 24.5     \\ \bottomrule
\end{tabular}
\caption{The performance of GraphERE with mix ratio $\beta$ in static event graphs. The average F1 scores of 10 experiments are used.}
\label{tab:2}
\end{table}

\subsection{Analysis for IE Graph and AMR Graph}
To explore the effects of IE Graph and AMR graph in static graphs on the model performance, we chose different mixed ratio $\beta$. When $\beta$ equals 0, the model uses only the IE graph, while when $\beta$ equals 1, the model uses only the AMR graph, and the detailed results are shown in Table \ref{tab:2}. In general, the effect of the model on the four event relations tends to increase as the percentage of the AMR graph grows, which we presume is because the AMR parser is capable of extracting more semantic relations and less noisy event arguments. However, when $\beta$ exceeds 0.8, the model shows a decrease in effectiveness on three tasks, validating the role of IE graphs that are more concerned with entity and entity relations. Therefore, we choose 0.8 as the $\beta$ value for the best overall performance.
\subsection{Data Scale Analysis}
The graph representation learning modules GAT and GCN are introduced in GraphERE, along with a multi-head attention mechanism. To analyze our method's data requirements, we conduct data scale experiments, and the results are shown in Figure \ref{fig:4}. It can be seen that our method consistently outperforms the baseline in the Subevent relation. In the three event relations of Coreference, Temporal and Causal, GraphERE performs worse than the baseline in extremely scarce data scenarios. When the data scale reaches 60\%, 40\%, and 40\%, respectively, our method overpasses the baseline and remains until 100\%. This indicates that GraphERE needs sufficient data to train the parameters in each module. In addition, GraphERE grows significantly faster than the baseline in each relation when the data scale is above 50\%, indicating that our method has better growth potential with additional data.
\section{Conclusion}
In this paper, we focus on the task of event-event relation extraction. We introduce the features of event argument and structure to obtain Graph-enhanced Event Embeddings. The GraphERE framework is constructed to exploit the information interaction between multiple event relations through Node Transformer and Task-specific Dynamic Graphs to achieve joint multiple event relation extraction.
The experimental results show that GraphERE significantly outperforms existing methods, which verifies its effectiveness. The results of the ablation analysis also validate the importance of each module.
In the future, we will explore open domain event relation extraction and distill our method to suit the needs of few-shot or zero-shot scenarios.
\section{Limitations}
The limitation of this paper's work lies mainly in using NLP tools, including AMR parsing and OpenIE extraction tools. These tools take a relatively long time to process documents. In particular, OpenIE uses the Stanford core NLP interface and does not support GPU acceleration, so processing each document is roughly between 5-10 seconds (8-core processor).
The good thing is that the static event graph generated for the same document does not change with training. Therefore, to speed up the training and testing process, we use AMR and OpenIE to preprocess all the training and testing sets documents and cache the generated event graphs.
\bibliography{anthology,custom}
\bibliographystyle{acl_natbib}




\end{document}